\documentclass[a4paper,fleqn]{cas-sc}
\usepackage{amssymb}
\usepackage{graphicx}

\usepackage{subcaption}
\usepackage{lipsum}
\usepackage{lscape}
\usepackage{amsmath}
\usepackage{multirow}
\usepackage[figuresright]{rotating}
\usepackage{lineno}
\usepackage{algorithm} 
\usepackage{algpseudocode} 
\usepackage{comment}
\usepackage{lineno}
\usepackage{nomencl}
\makenomenclature
\usepackage{calc}
\usepackage[T1]{fontenc}
\usepackage{relsize}
\usepackage{gensymb}
\usepackage{float}
\usepackage{svg}

\usepackage{lineno}
\usepackage{setspace}

\usepackage[sort,comma,authoryear,round]{natbib}

\setcitestyle{authoryear,open={(},close={)},citesep={;}} 

\hypersetup{
  colorlinks,
  citecolor=Violet,
  linkcolor=Red,
  urlcolor=Blue}

\newcolumntype{P}[1]{>{\centering\arraybackslash}p{#1}}

\def\tsc#1{\csdef{#1}{\textsc{\lowercase{#1}}\xspace}}
\tsc{WGM}
\tsc{QE}
\tsc{EP}
\tsc{PMS}
\tsc{BEC}
\tsc{DE}

\begin{document}
\let\WriteBookmarks\relax
\def\floatpagepagefraction{1}
\def\textpagefraction{.001}
\shorttitle{Foundation Model-Based Selective Harvesting}
\shortauthors{Zhu et~al.}
 
\title [mode = title]{Foundation Model-Based Apple Ripeness and Size Estimation for Selective Harvesting}

\author[1]{Keyi Zhu}\ead{zhukeyi1@msu.edu}
\author[2]{Jiajia Li}\ead{lijiajia@msu.edu}
\author[1]{Kaixiang Zhang}\ead{zhangk64@msu.edu}
\author[2]{Chaaran Arunachalam}\ead{arunach6@msu.edu}
\author[3]{Siddhartha Bhattacharya}\ead{bhatta70@msu.edu}
\author[4]{Renfu Lu}\ead{renfu.lu@usda.gov}
\author[1]{Zhaojian Li*}\ead{lizhaoj1@egr.msu.edu}

\address[1]{Department of Mechanical Engineering, Michigan State University, East Lansing, MI, USA}
\address[2]{Department of Electrical and Computer Engineering, Michigan State University, East Lansing, MI, USA}
\address[3]{Department of Computer Science and Engineering, Michigan State University, East Lansing, MI, USA}
\address[4]{United States Department of Agriculture Agricultural Research Service, East Lansing, MI, USA}

\address{* Zhaojian Li is the corresponding author}

\begin{abstract}
Harvesting is a critical task in the  tree fruit industry, demanding extensive manual labor and substantial costs, and exposing workers to potential hazards. Recent advances in automated harvesting offer a promising solution by enabling efficient, cost-effective, and ergonomic fruit picking within tight harvesting windows. However, existing harvesting technologies often indiscriminately harvest all visible and accessible fruits, including those that are unripe or undersized.
%, reducing revenue  and market value.  Evaluating fruit harvestability—-particularly ripeness and size-—prior to picking can significantly enhance profitability, especially for high-value varieties like Fuji apples.
%In the field, the \emph{harvest-ability} of a fruit can be evaluated by the fruit's appearance and size. Unlike the normal fruit detection problem, the difference between \emph{harvest-able} fruits and \emph{non-harvest-able} fruits is more subtle. Another challenge in conducting the in-field \emph{harvest-ability} evaluation is the loss of information caused by the canopy occlusion. In addition, the lack of corresponding dataset even boosts the challenges for this research topic.
This study introduces a novel foundation model-based framework for efficient apple ripeness and size estimation. %focus on the evaluation of \emph{harvest-ability} of Fuji apples. 
Specifically, we curated two public RGBD-based Fuji apple image datasets, integrating expanded annotations for ripeness ("Ripe" vs. "Unripe") based on fruit color and image capture dates. The resulting  comprehensive dataset, \textit{Fuji-Ripeness-Size Dataset}, includes 4,027 images and 16,257 annotated apples with ripeness and size labels.
Using Grounding-DINO, a language-model-based object detector,  we achieved robust apple detection and ripeness classification, outperforming other state-of-the-art models. Additionally, we developed and evaluated six size estimation algorithms, selecting the one with the lowest error and variation for optimal performance.
The \emph{Fuji-Ripeness-Size Dataset} and the apple detection and size estimation algorithms are made publicly available\footnote{The code and dataset is available at \url{https://github.com/zhukeyi-stan/Fuji\_Ripeness\_And\_Size\_Estimation}}, which provides valuable benchmarks for future studies in automated and selective harvesting.
\end{abstract}


\maketitle
\doublespacing

\section{Introduction}
\begin{figure}[htbp]
    \centering
    \includegraphics[width=0.45\textwidth]{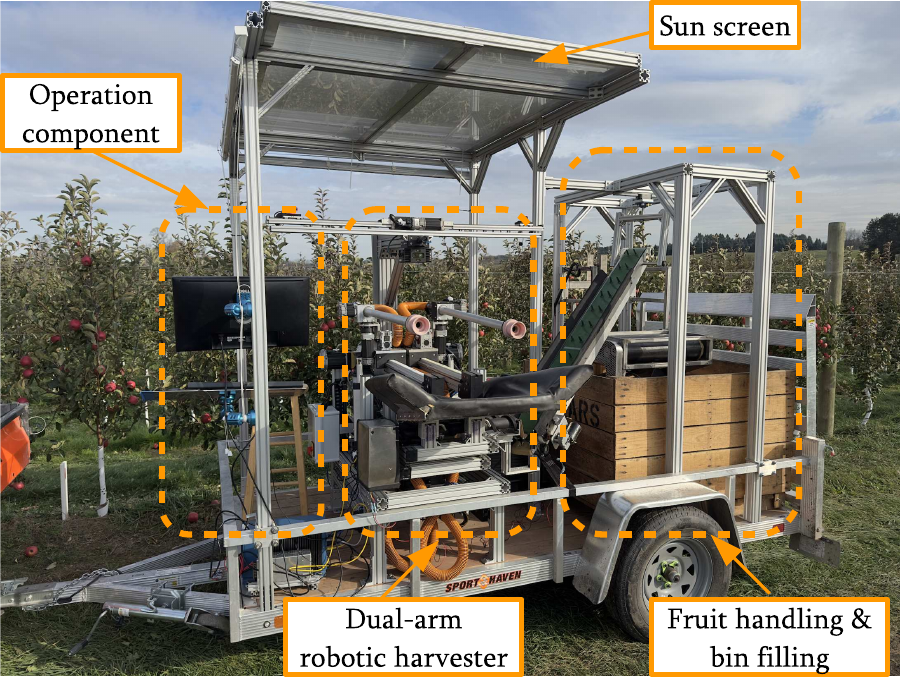}
    \caption{A photographic view of our apple harvesting robot. The system mainly consists of two 4 Degrees Of Freedom (DOF) robotic arms and a time-of-flight camera which is attached to the robot frame to localize the apples for picking.}
    \label{fig_system}
\end{figure}

Apple harvesting is a labor-intensive and costly operation. %The limited time window for harvesting ripe apples necessitates the rapid and efficient collection of apples within a short period, rendering the process labor-intensive and costly. 
With rising shortages and increasing costs of labor, the development of automated harvesting robots has attracted significant research interest in recent years~\citep{jia2020apple,au2023monash,yu2021lab,li2023multi}. Our group, for instance, has also been developing a vacuum-based apple harvesting robot in the past 5 years~\citep{Zhang2021MECH,Zhang2022IROS,Lu2022ASABE,Zhang2023CEA}, as shown in Fig.~\ref{fig_system}. These robots typically employ sensors such as RGB-D cameras alongside computer vision algorithms to detect and localize target fruits in the scene. Planning and control algorithms enable robotic arms to approach and harvest apples sequentially within the workspace. The integration of automated harvesting systems promises a transformative step forward in modern agriculture, addressing longstanding challenges in fruit production.

Fruit quality plays a crucial role in determining the value of harvested produce. Ripe fruits with appropriate size and shape are suitable for the fresh market, while unripe or misshapen fruits are relegated to processing, resulting in significantly lower market value. Experienced human pickers can efficiently assess ripeness and size during harvesting; however, the ability of automated systems to selectively harvest ripe and appropriately sized fruits has received limited attention.  %In the apple industry, fruit quality is a pivotal factor that influences market value. Ripe apples contribute to profitability, while unripe or damaged apples are often relegated to livestock feed or soil fertilization. For most apple varieties, harvesting occurs once during a specific period, typically determined by the grower's experience, ensuring that the majority of apples are ripe and of high quality. Subsequently, the apples are transported to packing houses equipped with specialized machinery and sensors for the evaluation of quality and ripeness.
This capability is particularly critical for premium apple varieties, such as \textit{Honeycrisp}, where higher profit margins justify multiple rounds of harvesting to maximize marketable yield. Achieving this requires an in-field assessment of apple ``harvestability''—evaluating whether an apple is ready for harvesting. Ideally, harvesting robots should incorporate algorithms capable of performing real-time ``harvestability'' evaluations. This paper reports our effort on developing robust computer vision algorithms to assess apple harvestability based on two key factors: ripeness and size.

%Recent advancements in computer vision have provided non-invasive tools for agriculture, effectively extracting features like color and shape using cost-efficient camera systems. While traditional methods, such as color thresholding~\citep{hamza2020design}, perform well in controlled settings, outdoor environments present challenges including occlusions, variable lighting, and complex backgrounds. Alternative techniques, such as K-means clustering and thresholding in non-RGB color spaces~\citep{sabzi2019automatic,goel2015fuzzy}, improve robustness under certain conditions but lack generalization in dynamic settings due to reliance on predefined parameters.

Ripeness and size estimations typically begin with an object detection algorithm to identify target fruits using 2D bounding boxes. To achieve this, various deep learning algorithms have been developed for robust object detection. Two-stage approaches, such as the R-CNN series~\citep{girshick2015fast,ren2015faster,he2017mask}, can achieve high precision by refining region proposals, while one-stage models like YOLO~\citep{wang2024yolov9}, RetinaNet~\citep{lin2017focal}, and SSD~\citep{liu2016ssd} provide faster inference. Transformer-based architectures, particularly Detection Transformers (DETR)~\citep{carion2020end} and its enhancements~\citep{zhu2020deformable,zhang2022dino}, have introduced self-attention mechanisms for global feature modeling, eliminating the need for post-processing. Grounding-DINO~\citep{liu2023grounding} further extends detection capabilities to open-set scenarios, integrating language prompts for tasks such as detecting ``red apples with smooth surfaces'' under challenging conditions like occlusions and varying lighting. These advancements hold immense potential for agricultural robotics, enabling accurate fruit detection, ripeness classification, and size estimation.

In-field ripeness estimation has received comparatively less attention than fruit detection. Traditional approaches often rely on image-based techniques, such as color-based fuzzy classification~\citep{hamza2020design} and SVM-based segmentation in various color spaces~\citep{pardede2019fruit,goel2015fuzzy}. While these methods perform well in controlled environments, their generalizability is limited under complex orchard conditions. The advent of deep learning has enabled the development of more adaptive models for ripeness estimation. For instance, convolutional neural networks (CNNs) have been applied to classify ripeness from fruit images~\citep{gunawan2021apple,saranya2022banana}, while transformer-based models~\citep{zhang2023attention} have achieved higher accuracies by leveraging attention mechanisms. However, most existing studies focus on simple backgrounds and isolated targets, which restrict their applicability in real-world orchards where fruits are often occluded by foliage and branches. To address these challenges, advanced models such as Mask R-CNN~\citep{ni2020deep} and DETR~\citep{xiao2021apple,hamza2021comparative,xiao2024apple} have been adapted to improve performance in complex scenarios. Additionally, model variants have been explored. For example, ~\citep{halstead2018fruit} added dual heads to Faster-RCNN for simultaneous fruit detection and ripeness prediction, while~\citep{lu2022canopy} integrated a lightweight Transformer into YOLOv4, enhancing detection and ripeness classification. 

Size estimation, on the other hand, has traditionally relied on manual tools like calipers, which are labor-intensive and impractical for large-scale operations. Reference object-based methods have also been explored~\citep{nenavath2024artificial,apolo2020deep}, but  they suffer from low computation efficiency, making them unsuitable for large-scale orchards. A more common approach involves  first estimating the size of the target fruit in 2D images using various techniques~\citep{freeman2023autonomous,mirbod2023tree,neupane2022orchard,blok2021image}, followed by transforming the pixel-based measurements into 3D dimensions. Recent advancements in fruit size estimation leverage 3D point clouds for improved accuracy, where depth cameras are used to capture 3D representations of fruit~\citep{gene2021field,miranda2023assessing,sapkota2024immature}. However,  the quality of  depth data from  off-the-shelf cameras often poses challenges, particularly in low-light conditions or when dealing with occluded fruits. To address these issues, deep learning methods have been developed. These include models that estimate both the visible and occluded portions of the fruit ~\citep{gene2023looking}, and end-to-end approaches for direct size estimation~\citep{ferrer2023simultaneous}, which enhance both robustness and accuracy in size estimation.

Despite these advancements, significant challenges remain in achieving robust apple ripeness and size estimation. %Ripeness estimation is actually determined by factors like firmness, acidity, and starch content~\citep{sabzi2022non}, which can only be assessed in sorting machines~\citep{pothula2023evaluation}, where the whole surface of the fruit can be evaluated with specialized sensors. 
%In-field ripeness estimation is complicated by the lack of standardized ripeness indicators across apple varieties. For instance, while most varieties exhibit a transition from green to red as they ripen, some, like \emph{Blondee}, turn yellow when ripe.
In outdoor settings, ripeness estimation relies heavily on color sensitivity, and varying lighting conditions introduce noise, complicating the analysis. Moreover, the inability to observe the entire surface of the apple, as compared to controlled rotation in sorting machines in the packinghouse, which further limits accuracy. Similarly, size estimation with RGB-D cameras faces challenges from noisy depth data and occlusions, particularly in dense canopies. Effective algorithms must account for these limitations while providing scalable solutions for large orchards. In addition, the lack of publicly available dataset on fruit ripeness and size estimation also hinders the advancement of this research area.

\begin{figure}[htbp]
    \centering
    \includegraphics[width=0.8\textwidth]{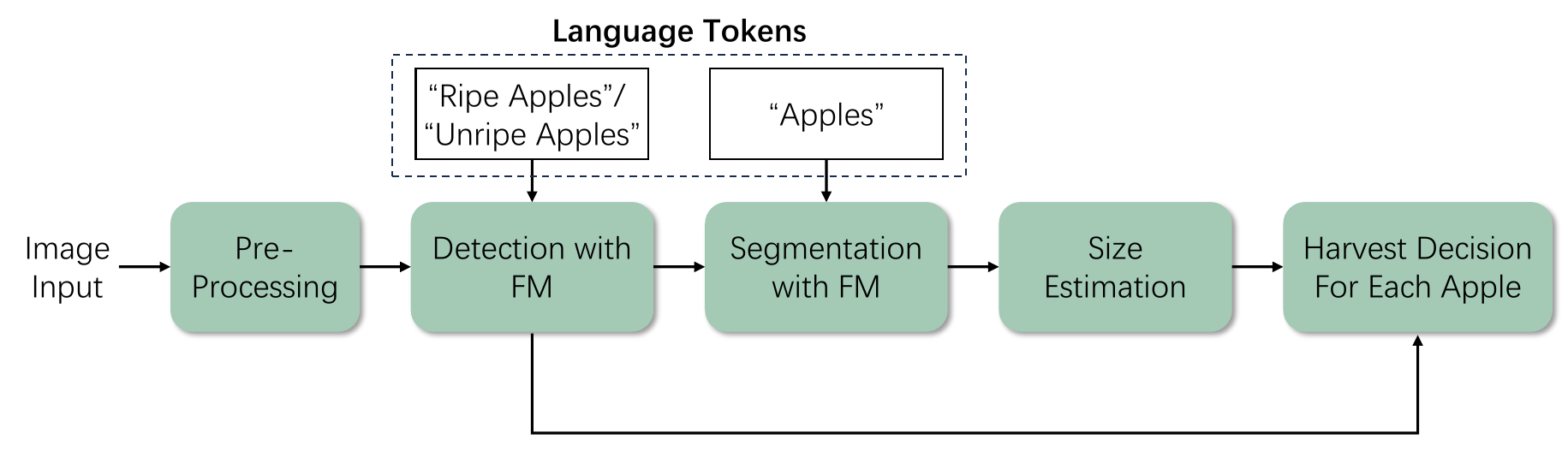}
    \caption{The procedure for in-field harvesting with evaluation of the "harvest-ability".}
    \label{workflow}
\end{figure}

To address the aforementioned challenges, in this article, we present a novel framework for accurate apple ripeness and size estimation, designed to enable automated selective harvesting.  As shown in Fig.~\ref{workflow}, our framework seamlessly integrates apple detection, ripeness classification, and size estimation of apples. Specifically, we curated and augmented two publicly available Fuji apple datasets to include ripeness estimates based on visual appearance and the time of the season. These datasets comprise  RGB-D images and corresponding size information of each apple. To achieve robust performance, we applied Grounding-DINO, a foundation-model-based detection framework, to detect apples and classify their ripeness.  This approach efficiently leverages dataset information and demonstrates strong adaptability to complex outdoor scenarios. Additionally, we proposed and evaluated six size estimation algorithms, with the best one achieving state-of-the-art performances. By combining ripeness classification and size estimation, our robotic system is equipped to make informed harvesting decisions for each apple. The key contributions of this paper are summarized as follows:
\begin{enumerate}
    \item \textbf{Dataset Augmentation:} We reprocessed two publicly available Fuji apple datasets containing RGB-D images and size information, to include ripeness annotations based on apple appearance. To our knowledge, this represents the first publicly available dataset with both ripeness and size annotations.
    \item \textbf{Model Development:} We employed Grounding-DINO, a foundation model-based network, to identify apples and classify their ripeness. This approach achieved higher mean average precision (mAP) compared to baseline detection models.
    \item \textbf{Size Estimation:} We developed six size estimation algorithms: three leveraging 2D image information and the other three utilizing 3D spatial data. A thorough comparison was conducted to evaluate their performance with box plots.
\end{enumerate}

\section{Materials and Methods}
In this section, we firstly introduce our dataset, detailing how we curated and augmented existing datasets. Next, we describe our apple detection and ripeness classification model. Then, we propose six size estimation algorithms.

\subsection{Fuji-Ripeness-Size Dataset} \label{dataset}

Although some works have studied fruit ripeness classification, their apple ripeness dataset were not made publicly available~\citep{sabzi2019automatic,zhang2023attention,lu2022canopy,sabzi2022non}. Two main factors make it difficult to create a comprehensive dataset for apple ripeness: (1) there is no universally accepted non-destructive standard for determining apple ripeness, and (2) ripening characteristics vary greatly among different apple varieties. Nonetheless, using surface color and image capture date within a single variety offers a practical approach to generating an apple ripeness dataset.
Thus, in this study, we address the lack of public apple ripeness datasets by curating two existing Fuji apple datasets, described below. While these datasets were originally designed for apple detection and size measurement, we augmented them with ripeness labels.

\begin{figure}[htbp]
	\centering
	\begin{subfigure}[b]{0.6\textwidth}
		\centering
		\includegraphics[width=1\textwidth]{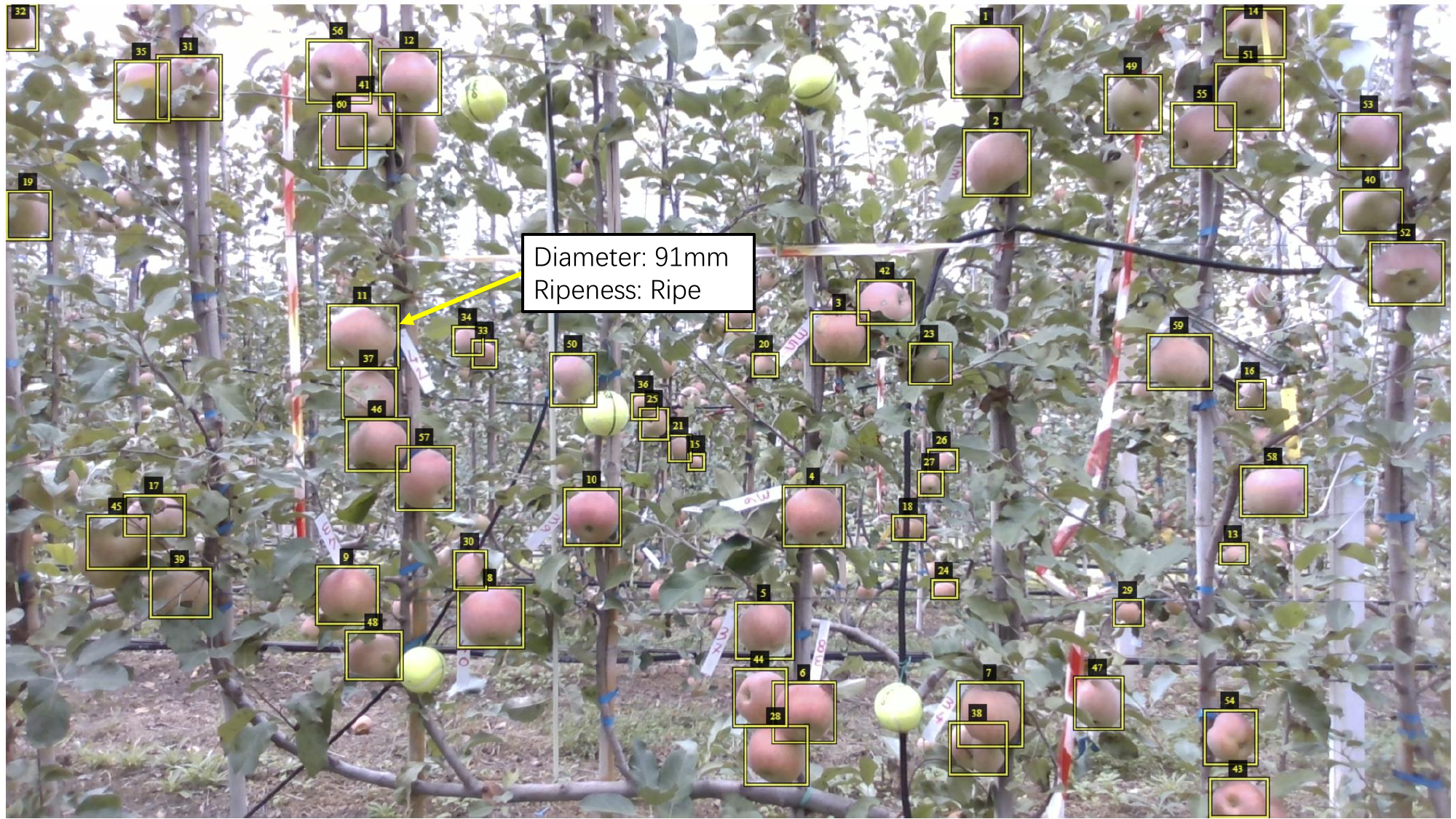}
		\caption{}
		\label{fig_unibo_sample}
	\end{subfigure}
	\hfill
    % \hspace{15pt}
	\begin{subfigure}[b]{0.34\textwidth}
		\centering
		\includegraphics[width=1\textwidth]{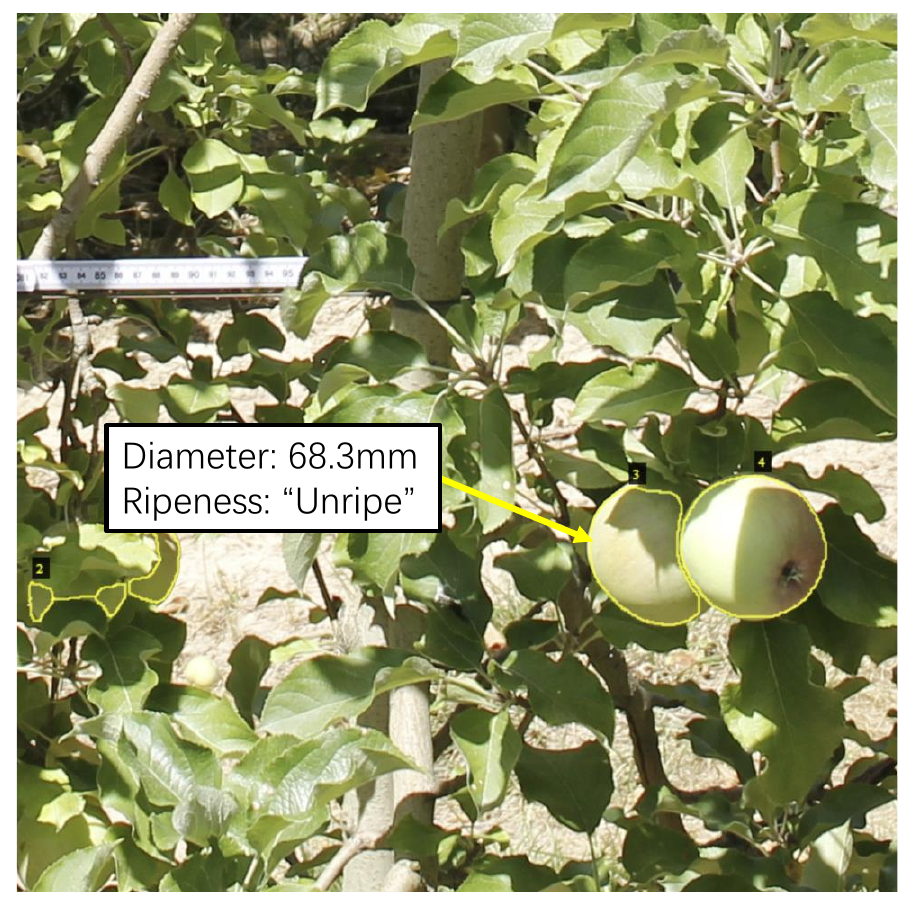}
		\caption{}
		\label{fig_udl_sample}
	\end{subfigure}
    \caption{The sample images for each dataset after re-annotation: (a) OpenAcces\_RGBD\_Apple\_Dataset; and (b) AmodalAppleSize\_RGB-D Dataset.}
    \label{fig_datasets}
\end{figure}

\textbf{OpenAcces\_RGBD\_Apple\_Dataset} \citep{bortolotti2024openacces_rgbd_apple_dataset} was gathered in 2022 at the University of Bologna, Italy, in a three-year-old orchard with Fuji apples trained as ``planar cordons''. This training style helps reduce occlusions (see Fig.~\ref{fig_unibo_sample} for a sample image). The researchers tracked 24 apples on two trees, measuring their maximum equatorial diameters with a digital caliper at several stages. They then captured images from multiple viewpoints (top, bottom, and full canopy) and at two distances from the camera (1.0 m and 1.5 m). An Intel RealSense D435i camera was used throughout, with settings fixed to achieve optimal depth precision (848×480) and RGB resolution (1920×1080). Internal camera functions handled alignment between depth and RGB images. In total, this dataset includes 102 RGB-D images under various lighting conditions, containing 922 apples with diameters ranging from 40 mm to 95 mm. Each apple was annotated with a bounding box.

\textbf{AmodalAppleSize\_RGB-D Dataset} \citep{gene2024amodalapplesize_rgb} was collected in a Fuji apple orchard in Agramunt, Catalonia, Spain, where trees were trained in a tall spindle system. The researchers photographed the canopy using a Canon EOS 60 DSLR camera (18 MP CMOS APS-C sensor, EFS 24,mm f/2.8 STM lens) from multiple angles, ensuring more than 75\% overlap among consecutive images (see Fig.~\ref{fig_udl_sample} for a sample image). They then applied Structure-from-Motion (SfM) and Multi-View Stereo (MVS) in \emph{Agisoft Professional Metashape software} to build a 3D point cloud of the canopy. Apples were tagged, and their sizes were recorded using a Vernier caliper. Data were collected on two dates—October 3, 2018, when most apples were ripe (BBCH85), and July 16, 2020, when most apples were unripe (BBCH77). The researchers cropped raw images to 1300×1300 pixels and generated depth maps aligned with each RGB image. They provided two segmentation masks for each apple: a modal mask (visible pixels) and an amodal mask (visible plus occluded pixels). Altogether, this dataset contains 3925 RGB-D images and 15335 labeled apples.

\textbf{Fuji-Ripeness-Size Dataset} was then curated by  augmenting the two datasets above. Note that the \textit{OpenAccess\_RGBD\_Apple} dataset and the \emph{AmodalAppleSize\_RGB-D} dataset differ in several key aspects. The \emph{OpenAccess\_RGBD\_Apple} dataset includes only bounding-box annotations, whereas the \emph{AmodalAppleSize\_RGB-D} dataset provides both modal and amodal annotations. Additionally, the \emph{OpenAccess\_RGBD\_Apple} dataset features canopy-wide images containing many apples, while the \emph{AmodalAppleSize\_RGB-D} dataset crops its images, resulting in fewer apples per frame. Apples in the \emph{OpenAccess\_RGBD\_Apple} dataset experience less occlusion due to the orchard’s training system, whereas those in the \emph{AmodalAppleSize\_RGB-D} dataset are more frequently obscured by branches and leaves. The \emph{AmodalAppleSize\_RGB-D} dataset also provides higher-precision depth maps than the \emph{OpenAccess\_RGBD\_Apple} dataset, reflecting differences in depth capture and processing methods. Furthermore, the \emph{OpenAccess\_RGBD\_Apple} dataset exhibits greater variability, having been collected over ten different days, while the \emph{AmodalAppleSize\_RGB-D} dataset was gathered on only two collection dates.

\begin{comment}
These two datasets vary in several important ways: 
\begin{enumerate} 
\item The \emph{OpenAcces\_RGBD\_Apple\_Dataset} includes only bounding-box annotations, while the \emph{AmodalAppleSize\_}\emph{RGB-D Dataset} provides both modal and amodal annotations. 
\item The \emph{OpenAcces\_RGBD\_Apple\_Dataset} features canopy-wide images containing many apples, whereas the \emph{AmodalAppleSize\_RGB-D Dataset} crops its images, resulting in fewer apples per image. 
\item In the \emph{OpenAcces\_RGBD\_Apple\_Dataset}, apples experience less occlusion because of the orchard’s training system; in contrast, apples in the \emph{AmodalAppleSize\_RGB-D Dataset} are more commonly obscured by branches and leaves. 
\item The \emph{AmodalAppleSize\_RGB-D Dataset} provides higher-precision depth maps than the \emph{OpenAcces\_RGBD\_} \emph{Apple\_Dataset}, reflecting differences in how depth was captured and processed. 
\item The \emph{OpenAcces\_RGBD\_Apple\_Dataset} has more variability overall, having been collected over 10 different days; the \emph{AmodalAppleSize\_RGB-D Dataset} comprises only two collection dates. \end{enumerate}
\end{comment}

\begin{figure}[htbp]
	\centering
	\begin{subfigure}[b]{0.35\textwidth}
		\centering
		\includegraphics[width=1\textwidth]{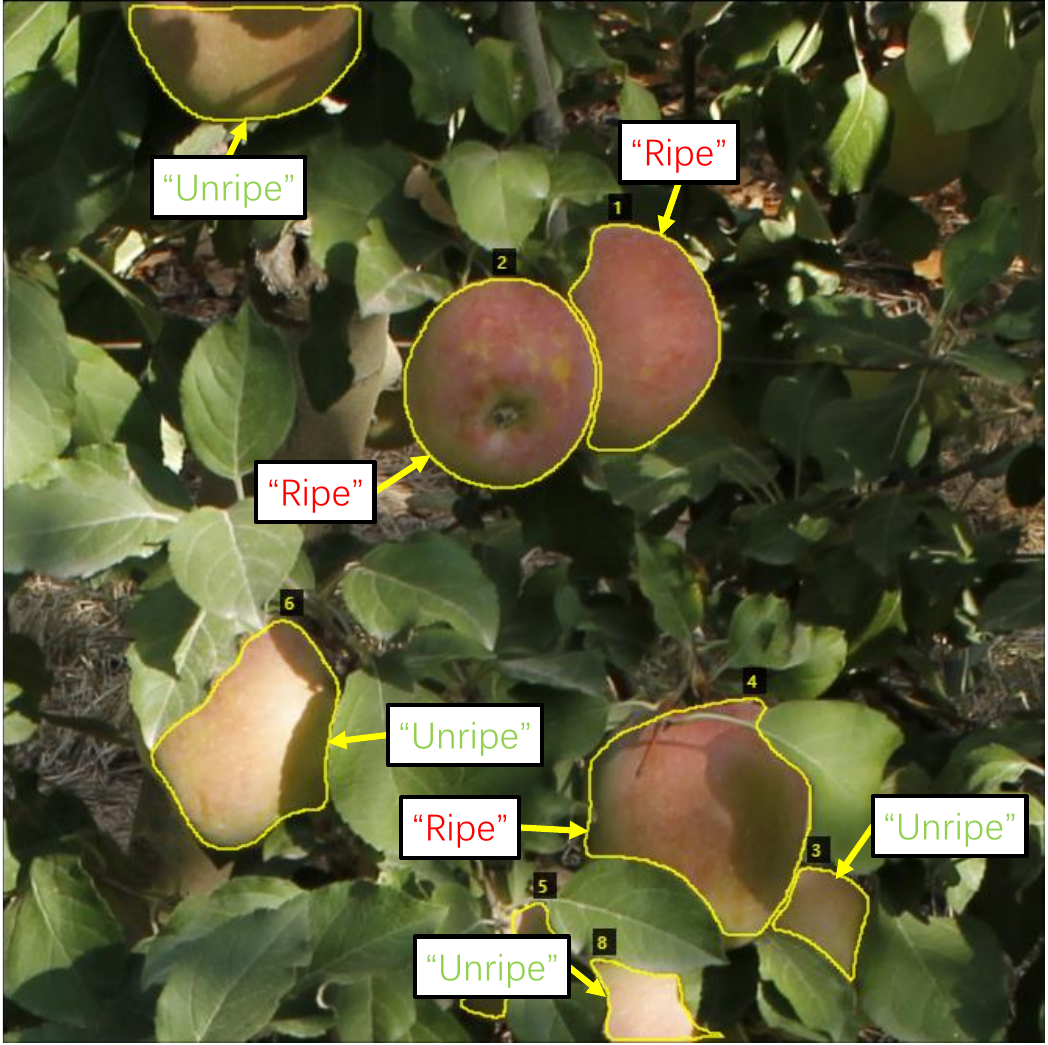}
		\caption{}
		\label{fig_ripeness_1}
	\end{subfigure}
	% \hfill
    \hspace{10pt}
	\begin{subfigure}[b]{0.35\textwidth}
		\centering
		\includegraphics[width=1\textwidth]{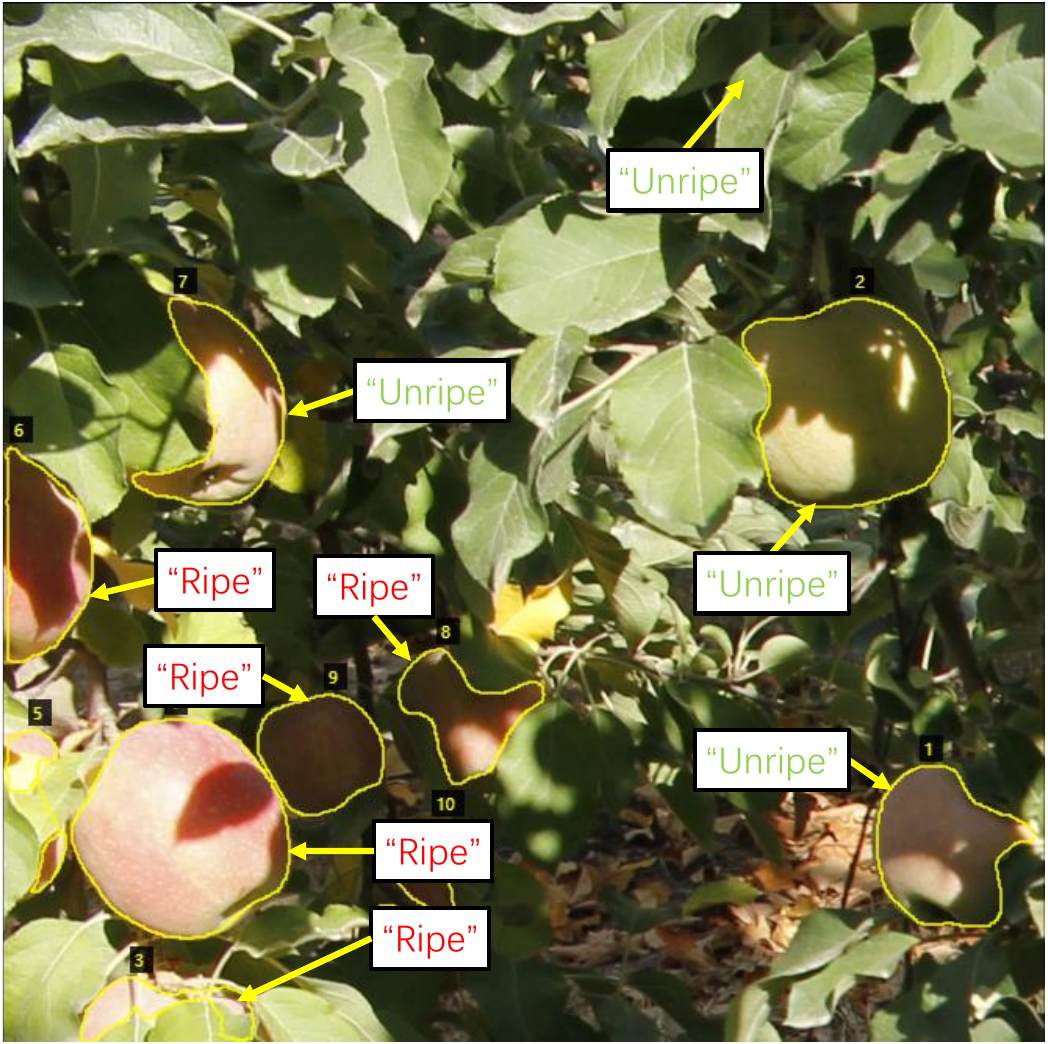}
		\caption{}
		\label{fig_ripeness_2}
	\end{subfigure}
\caption{Examples of ripeness labeling. The image collection date served as a preliminary reference; for instance, images captured in July were assumed to predominantly contain unripe apples. Subsequently, the ripeness of each apple was determined based on the percentage of its surface area covered by red regions.}
\label{fig_ripeness}
\end{figure}

Despite these differences, both datasets include Fuji apples at multiple ripening stages as well as size measurements for some apples. To enable  network training for simultaneous ripeness and size estimation, we augmented the above two datasets by assigning a ripeness label to each labeled apple based on its color and capture date, designating apples that appeared mostly red as ''Ripe'' and those that appeared mostly green as ''Unripe''; see Fig.~\ref{fig_ripeness} for a few samples. We annotated these apples using the open-source VIA tool~\citep{dutta2019vgg}. Because the \emph{AmodalAppleSize\_RGB-D} dataset provides pixel-wise segmentation masks whereas the \emph{OpenAcces\_RGBD\_Apple} dataset includes only bounding boxes, we converted the pixel-wise annotations in the former to bounding boxes for consistency. This process yielded a Fuji apple dataset with both ripeness labels and size data, which we then used to train and evaluate our ripeness model and to develop our size estimation algorithm.

\subsection{Apple Detection and Ripeness Classification with Grounding-DINO}

Recent advances in deep learning have led to significant breakthroughs in object detection. Models such as Faster-RCNN~\citep{ren2015faster} and the YOLO series~\citep{wang2024yolov9} can be trained on images and corresponding labels, which include both the position of target objects and their class information. Compared with traditional algorithms based on hand-crafted features, these deep-learning-based methods offer superior adaptability and generalization. They have been widely adopted for fruit detection and have shown promising results~\citep{ChuPRL2021, Chu2023arXiv, FuCEA2020, bargoti2017deep}.

However, most of these methods are closed-set detectors: they assume that all target classes are fully known during both training and inference, which limits their ability to handle previously unseen classes. With the rapid progress of natural language processing (NLP), researchers have integrated \emph{Transformers}~\citep{vaswani2017attention}, or attention-based mechanisms, into object detection architectures. This integration has given rise to open-set detectors, which use language processing components to go beyond fixed, predefined class sets. These open-set detectors not only excel at recognizing classes learned during training but can also detect new classes with the help of language models.

In this work, we employ Grounding-DINO~\citep{liu2023grounding}—one of the most advanced open-set detectors—to perform apple detection and ripeness classification. Grounding-DINO is a vision foundation model that combines DINO~\citep{zhang2022dino}, a Transformer-based closed-set detector, with Grounded Language-Image Pre-training (GLIP)~\citep{li2022grounded}. It can be trained on standard image datasets (with bounding boxes) while leveraging class names to boost its object recognition ability through language modeling. Unlike common datasets such as VOC or COCO~\citep{lin2014microsoft, everingham2010pascal}, which contain objects that differ markedly from one another, our task involves distinguishing only two very similar classes—``ripe apples'' and ``unripe apples''. Integrating language features helps the model better capture subtle differences between these two classes.

\begin{figure}[htbp]
    \centering
    \includegraphics[width=0.98\textwidth]{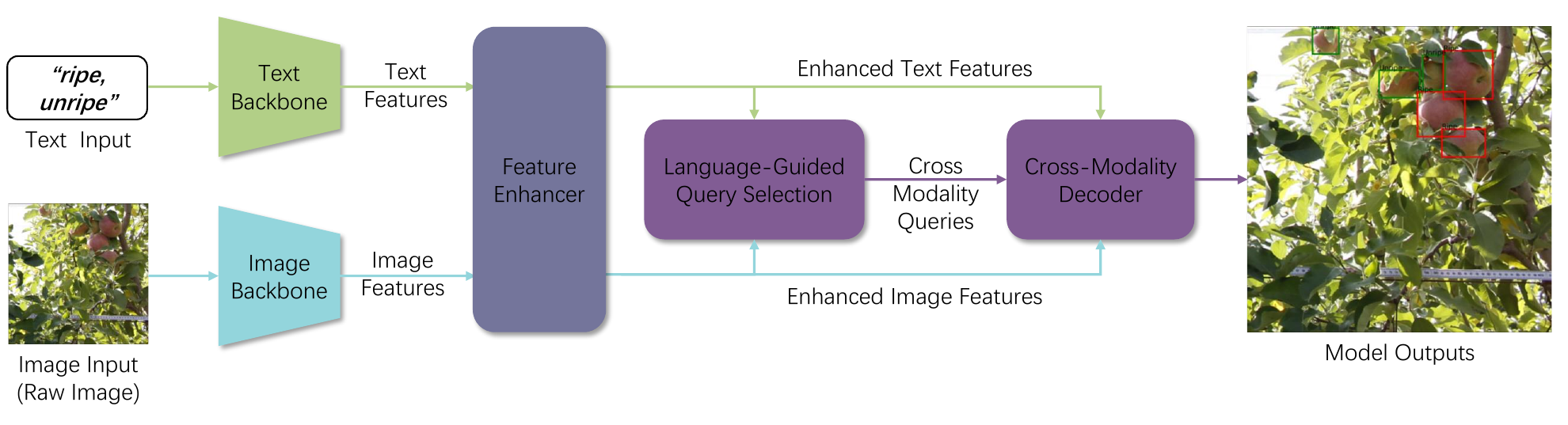}
    \caption{The network structure for the Grounding-DINO, where text prompt is also included as an input. The image and the text are joint inputs to the model at the same time, after feature extraction and cross feature enhancement, the model is able to infer from the cross-modality information and detect the target region specified by the text.}
    \label{GDINO_struct}
\end{figure}

Figure~\ref{GDINO_struct} outlines the Grounding-DINO architecture. The model receives a text prompt and an image as input and outputs the bounding box of the target object described by the text. Two separate backbones process the text and the image, extracting preliminary features from each. These features are refined through a feature enhancer layer consisting of self-attention and cross-attention mechanisms, which produce enhanced text and image features. The system then performs language-guided query selection and cross-modality decoding. Here, \emph{queries} are essentially encoded predictions on the enhanced image features, guided by the enhanced text features; a cross-modality decoder uses cross-attention to refine these queries further.

During training, Grounding-DINO employs L-1 and Generalized Intersection over Union (GIoU) losses to refine bounding-box coordinates. Following GLIP~\citep{li2022grounded}, it also adopts a contrastive loss between predicted objects and language tokens for classification. The overall loss function can be written as: $\mathcal{L}=\mathcal{L}_1+\mathcal{L}_{GIOU}+\mathcal{L}_{Cont},$ where $\mathcal{L}_1$, $\mathcal{L}_{GIOU}$, and $\mathcal{L}_{Cont}$ are the L1 loss, GIoU loss, and contrastive loss, respectively. 

To speed up training and improve performance, we use transfer learning, a common practice in deep-learning research. Specifically, we begin with a Grounding-DINO model pretrained on large-scale object detection datasets (e.g., COCO, O365, OpenImage, ODinW-35, RefCOCO~\citep{lin2014microsoft, shao2019objects365, li2022elevater, kazemzadeh2014referitgame, kuznetsova2020open}), then fine-tune it on our apple ripeness dataset. This approach harnesses the broad visual knowledge the model gains from extensive pre-training, enabling more accurate ripeness classification and detection in our specific application.

\subsection{Size Estimation}\label{sec_sizing}
%Fruit size estimation has been studied extensively for many years. Vision-based approaches offer a contactless way to obtain size information from camera images, which is far more efficient than manual measurement with calipers. 
Existing studies on fruit size estimation can be broadly grouped according to the type of data used. The first category relies on 2D images~\citep{nenavath2024artificial,mirbod2023tree,freeman2023autonomous}, where some methods compare detected fruits against a known reference object in the scene, while others estimate fruit size by some image processing algorithms. The estimated size in pixel units is then converted into real-world measurements (e.g., centimeters) via target-object distance and camera parameters. The second category uses 3D point clouds~\citep{tsoulias2023situ,gene2021field}, typically generated by sensors such as LiDAR or RGB-D cameras. Various computational methods are then applied to these 3D data to estimate fruit size.

In this article, we first segment apple pixels within the detected bounding boxes. After outlier removal, a filtering algorithm to get rid of noises in depth images, we implement several size estimation algorithms and compare their results to determine which method provides the most accurate size measurements.

\subsubsection{Mask Generation}
To remove irrelevant information from the bounding boxes, we apply a segmentation algorithm to isolate only apple pixels. Specifically, we use the Segment-Anything Model (SAM)~\citep{kirillov2023segany}, which is trained on a large-scale, web-based dataset of diverse objects. Rather than fine-tuning the model on our limited dataset, we directly prompt the model with the word ''apple'' to generate the segmentation masks. Because the apples were already localized and classified by ripeness in the previous step, using the general prompt ''apple'' (as opposed to ''ripe apple'' or ''red apple'') achieves good segmentation performance.
Once pixel-wise segmentation masks are generated for each bounding box, these masks are used in subsequent steps to estimate the size of each detected apple.

\begin{figure}[htbp]
	\centering
	\begin{subfigure}[b]{0.35\textwidth}
		\centering
		\includegraphics[width=1\textwidth]{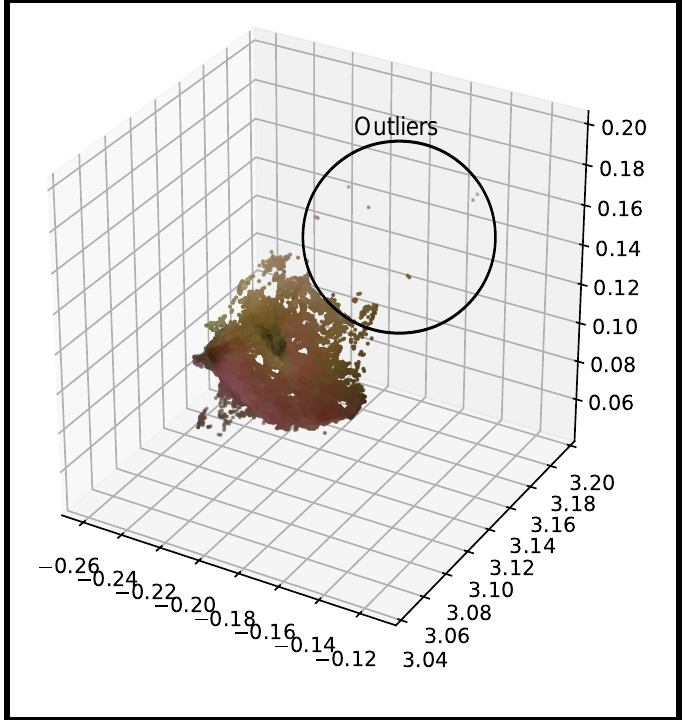}
		\caption{}
		\label{fig_unfiltered_pc}
	\end{subfigure}
	% \hfill
    \hspace{10pt}
	\begin{subfigure}[b]{0.35\textwidth}
		\centering
		\includegraphics[width=1\textwidth]{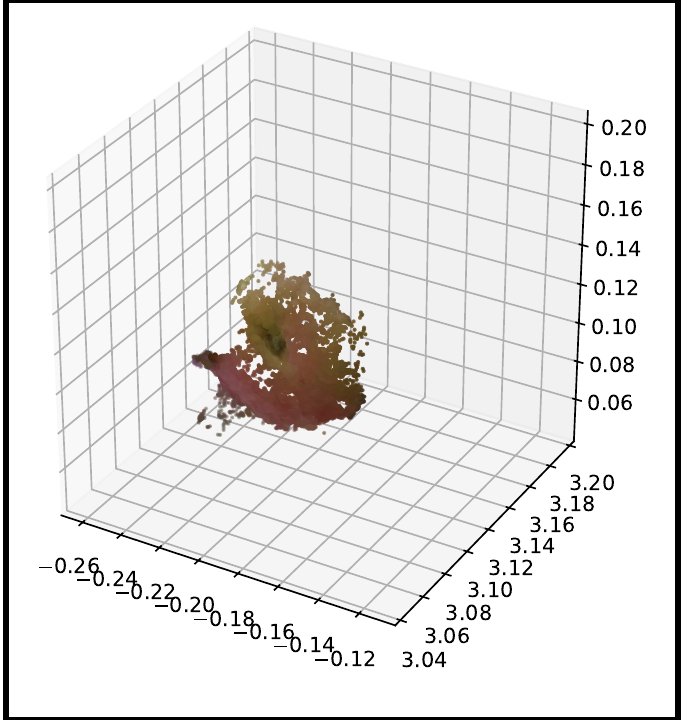}
		\caption{}
		\label{fig_filtered_pc}
	\end{subfigure}
\caption{Illustration of outlier removal in point cloud of an apple: (a) before and (b) after outlier removal. Although the outliers are usually very few compared to the inliers, they sometimes would bring significant error to the size estimation. Thus, the removal of outliers is very important in outdoor fruit sizing.}
\label{fig_outlier_removal}
\end{figure}

\subsubsection{Outlier Removal}\label{sec_outlier_removal}
Depth images often contain noisy measurements, so we remove outliers before proceeding with size estimation. Specifically, for each apple mask, we sort the corresponding depth pixels by their depth values. We then keep only those pixels falling within a certain percentile range (e.g., 10\% to 90\%). Figure~\ref{fig_outlier_removal} shows an example where the range is set to 20\%--80\%. This process effectively discards aberrant depth points while preserving the overall shape of the apple, thereby reducing errors during size estimation.

\subsubsection{2D-Based Estimation}
In 2D-based algorithms, the size of the apple is first estimated in pixel units before being converted to millimeters. Let $d_p$ be the pixel-level measurement, and $d_r$ be the real-world (e.g., millimeter) measurement. The conversion is given by $d_r=\frac{d_p}{f}\overline{z},$ (Fig.~\ref{fig_pc_gen}) where $f$ is the camera’s focal length, and $\overline{z}$ is the average distance between the camera and the apple (obtained by averaging all depth values within the mask, after outlier removal as described in Section~\ref{sec_outlier_removal}). Since these methods rely on 2D image measurements, outlier removal is only applied when calculating $\overline{z}$, in order to retain as many useful RGB pixels as possible.

In this article, we consider three 2D-based size estimation algorithms:
\begin{enumerate}
\item The size of the Bounding Box (\textbf{2D-BBox}, Fig.~\ref{fig_2d_bbox}): Let $h$ be the height of the bounding box, $w$ be the width of the bounding box, the pixel-wise diameter of the apple is estimated as $d_p=max(h,w)$.
\item Largest Segment within the mask (\textbf{2D-LSeg}, Fig.~\ref{fig_2d_lseg}): Denote the mask as $\textbf{M}=\{m_i\}$, where $m_i=(u_i,v_i)$ is the individual pixel within the mask. In this algorithm, the diameter of the apple is estimated as $d_p=\max\limits_{i,j}||m_i-m_j||_2$.
\item Hough Transform (\textbf{2D-HT}, Fig.~\ref{fig_2d_ht}): The Hough Transform is a well-known feature extraction method used here for circle fitting. We apply the transform to each apple mask to estimate its pixel-level diameter, $d_p=HT(mask)$.
\end{enumerate}

\subsubsection{3D-Based Estimation}
\begin{figure}[htbp]
    \centering
    \includegraphics[width=0.7\textwidth]{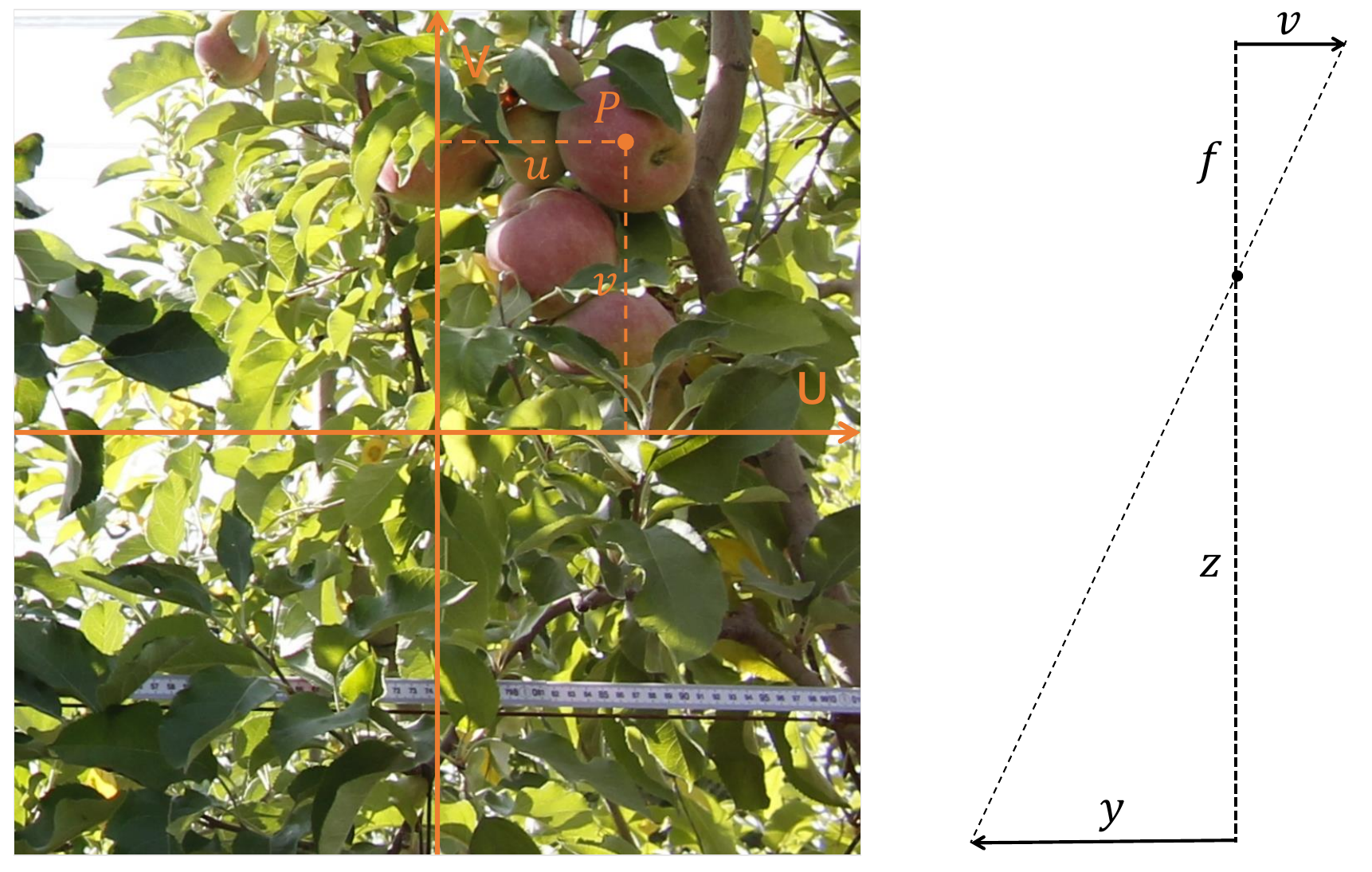}
    \caption{The conversion example from 2D image to 3D coordinate. Consider a point $P$ on the left image with pixel coordinate $(u,v)$ and according to the triangular geometry in the right image,  $v$ can be converted to $y$ with the depth value $z$ of point $P$ from the depth channel and the camera's focal length $f$. The transformation from $u$ to $x$ can be done analogously. Then $(x,y,z)$ is determined as the 3D position of the point $P$.}
    \label{fig_pc_gen}
\end{figure}

In 3D-based algorithms, the apple pixels are first converted into point-cloud data. Since the RGB and depth images in our datasets are pre-registered to the same resolution, each pixel $(u,v)$ in the RGB image corresponds to a pixel in the depth image. Suppose $z$ is the depth value at $(u,v)$, and $f$ is the camera’s focal length. The transformation to 3D coordinates is given by: $x=\frac{u}{f}z,\,y=\frac{v}{f}z.$ Following this mapping (illustrated in Fig.~\ref{fig_pc_gen}), each pixel in the segmentation mask becomes a 3D point, yielding a point cloud of the apple.

\begin{figure}[!t]
    \centering
	\begin{subfigure}[b]{0.27\textwidth}
		\centering
		\includegraphics[width=1\textwidth]{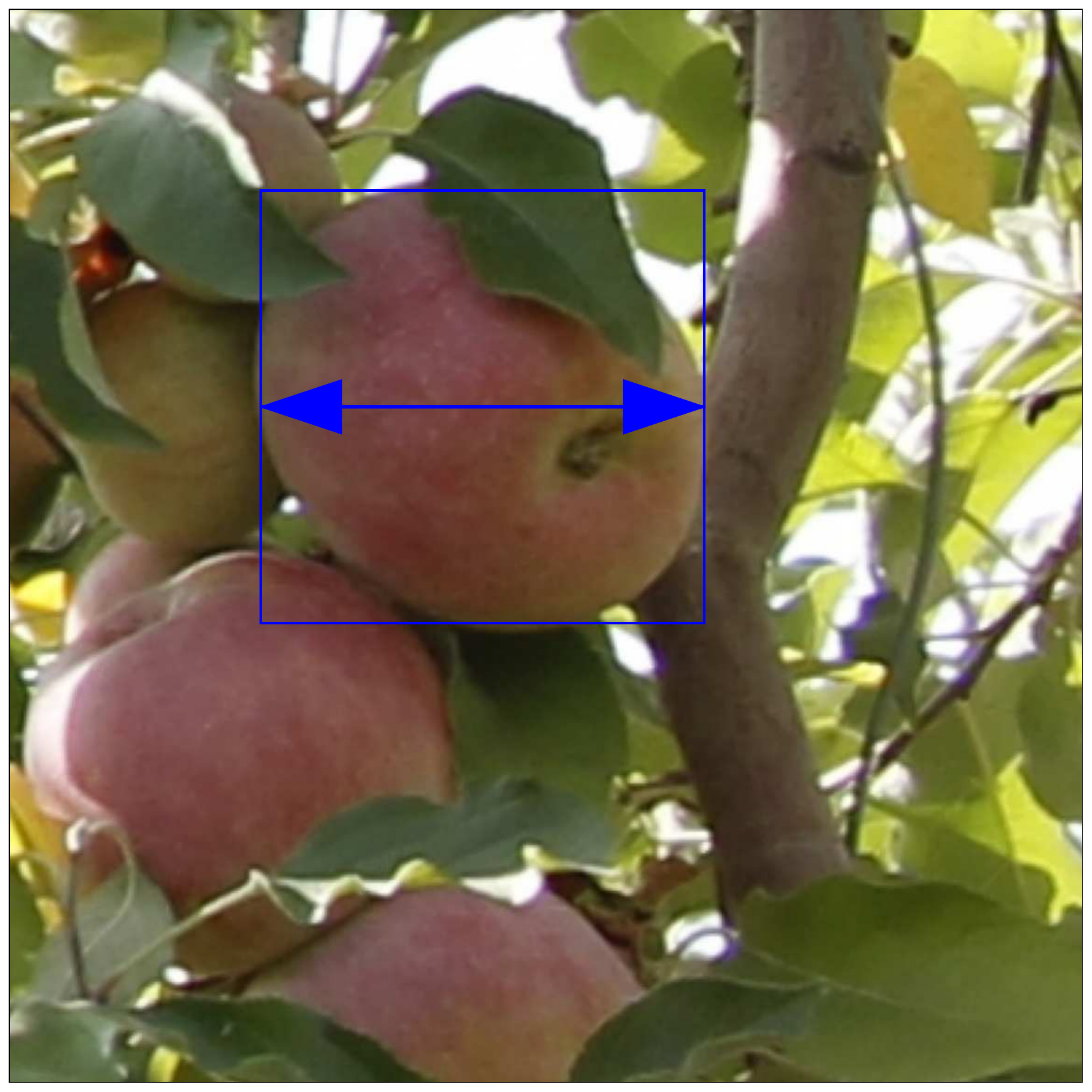}
		\caption{2D-BBox}
		\label{fig_2d_bbox}
	\end{subfigure}
	% \hfill
    % \hspace{15pt}
	\begin{subfigure}[b]{0.27\textwidth}
		\centering
		\includegraphics[width=1\textwidth]{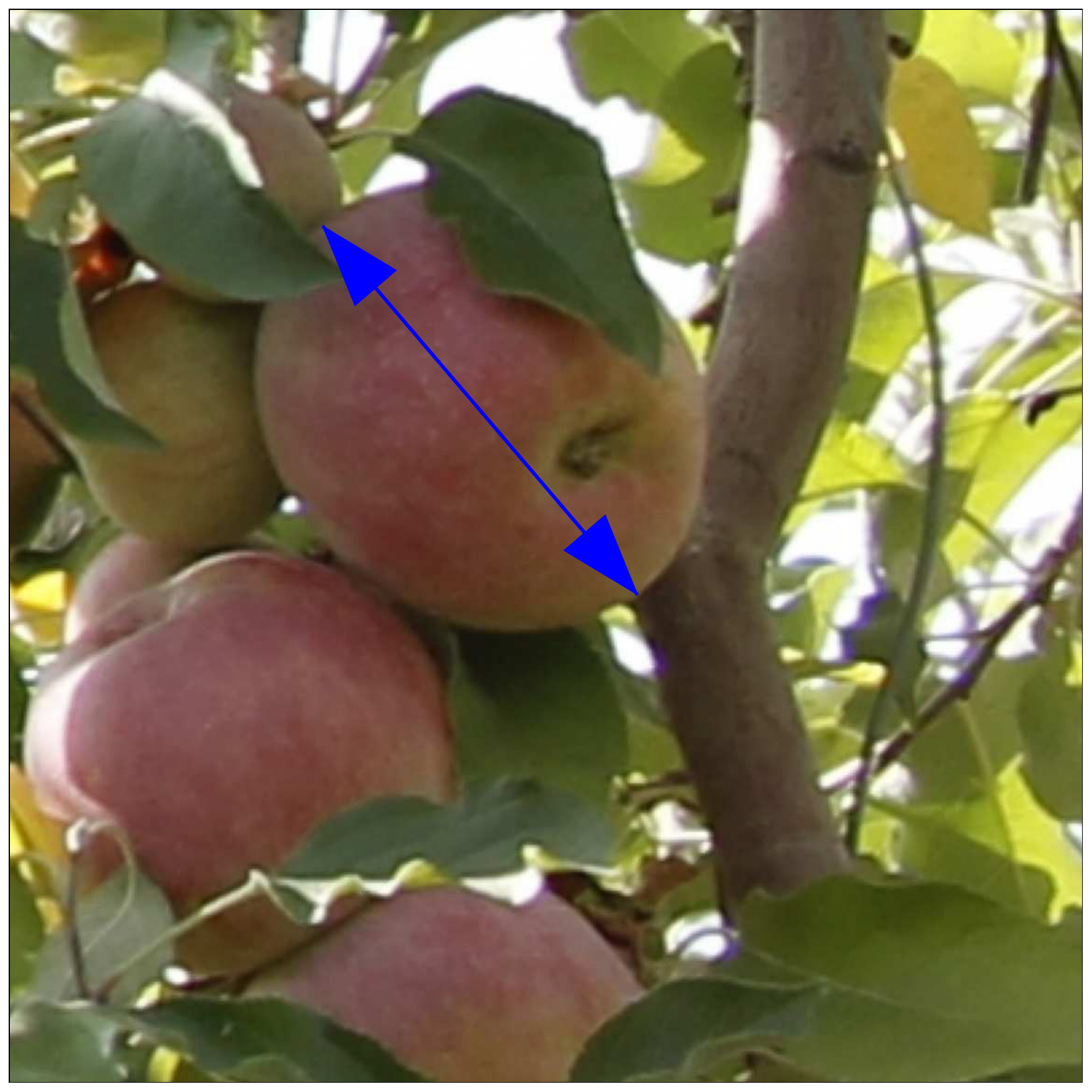}
		\caption{2D-LSeg}
		\label{fig_2d_lseg}
	\end{subfigure}
    % \hspace{15pt}
	\begin{subfigure}[b]{0.27\textwidth}
		\centering
		\includegraphics[width=1\textwidth]{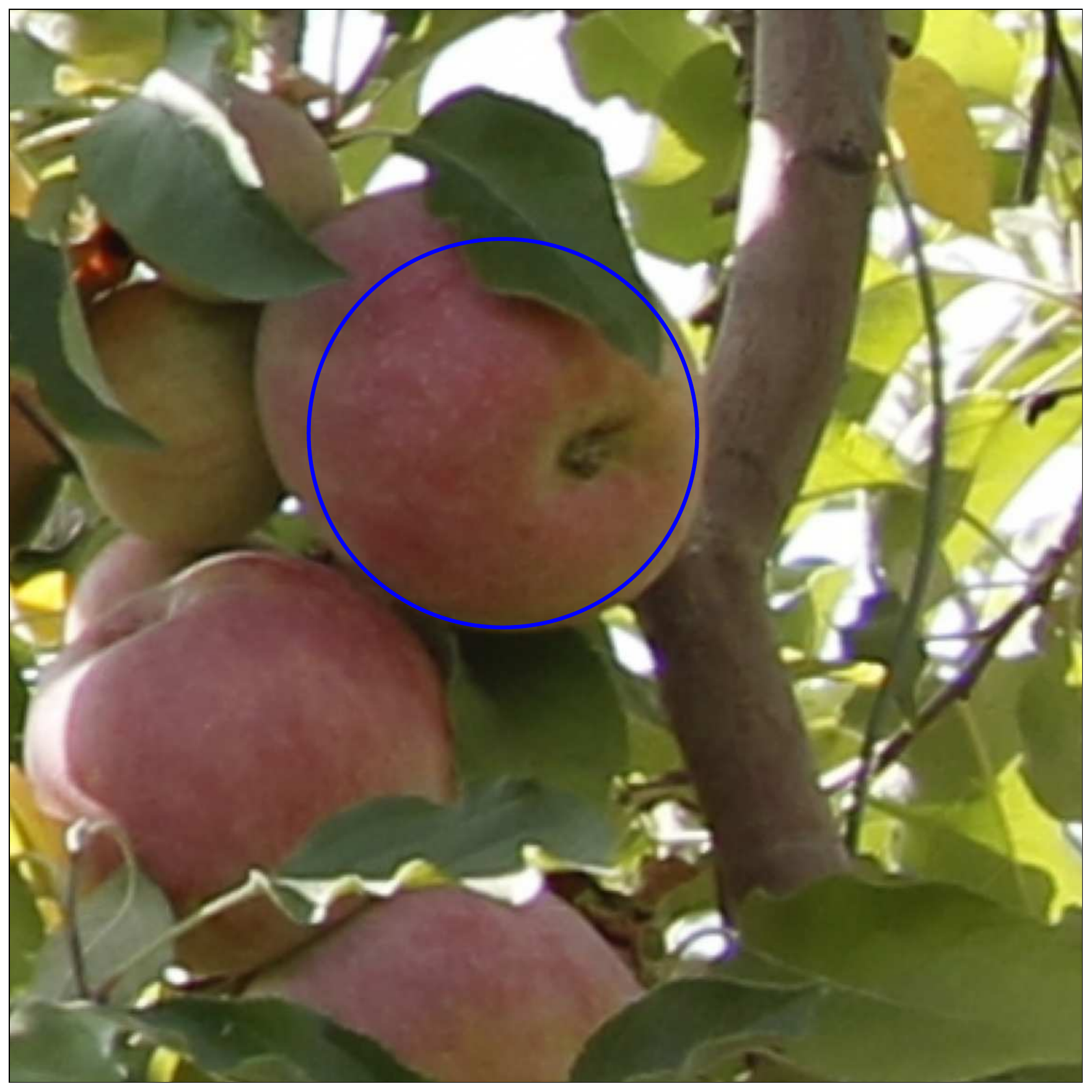}
		\caption{2D-HT}
		\label{fig_2d_ht}
	\end{subfigure}
    % \hspace{15pt}
    
	\begin{subfigure}[b]{0.27\textwidth}
		\centering
		\includegraphics[width=1\textwidth]{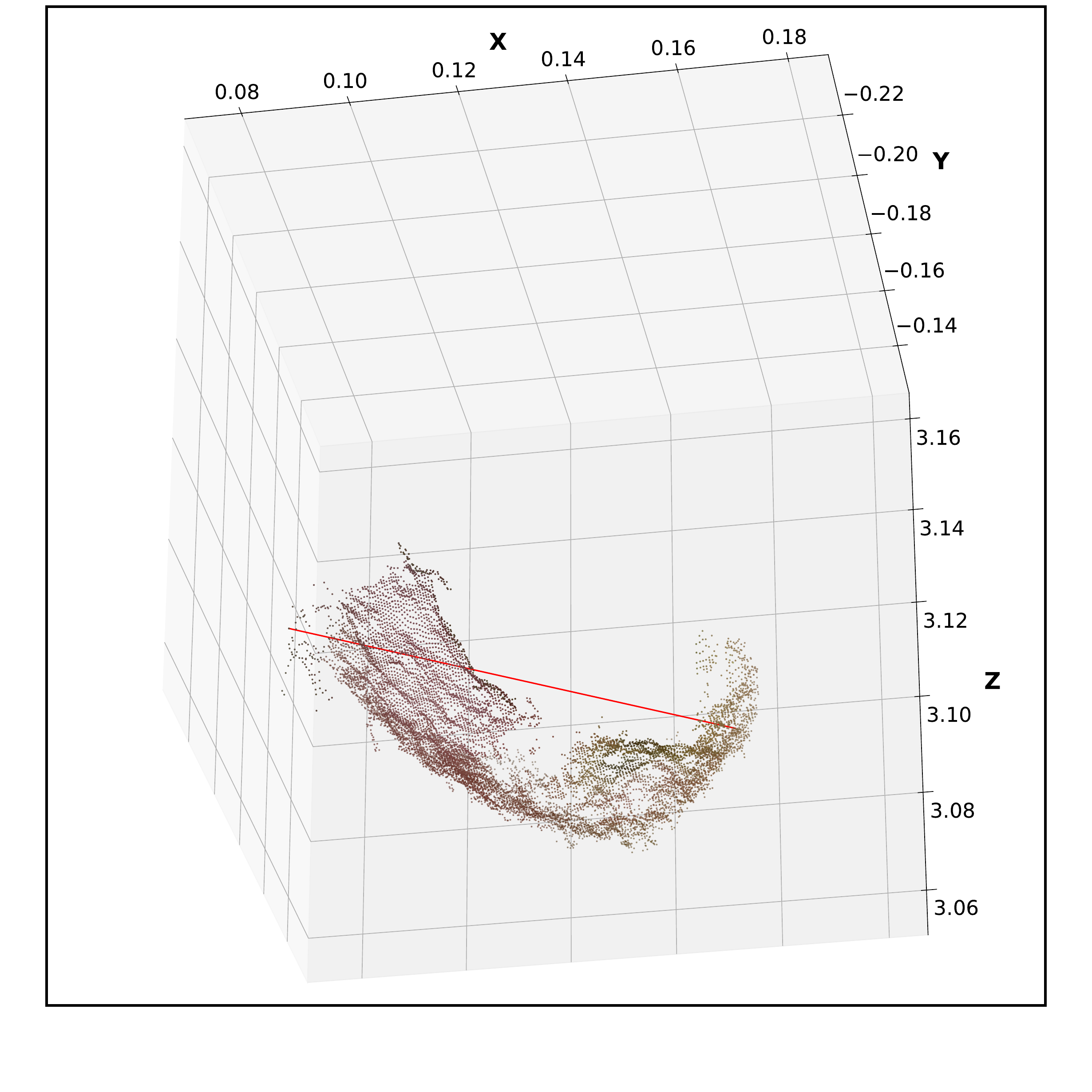}
            % \vspace{-30pt}
		\caption{3D-LSeg.}
		\label{fig_3d_lseg}
	\end{subfigure}
    % \hspace{15pt}
	\begin{subfigure}[b]{0.27\textwidth}
		\centering
		\includegraphics[width=1\textwidth]{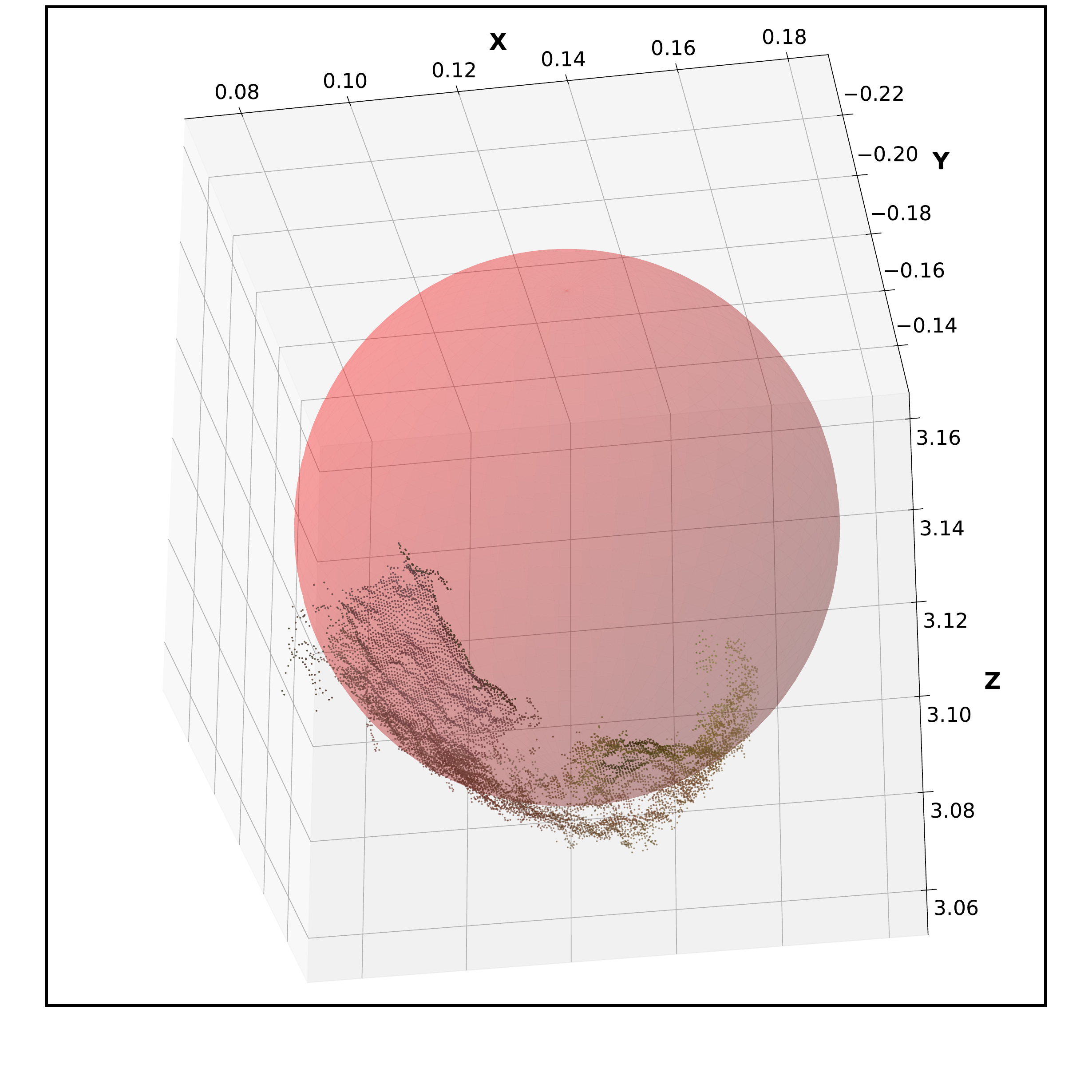}
            % \vspace{-30pt}
		\caption{3D-LSq}
		\label{fig_3d_lsq}
	\end{subfigure}
    % \hspace{15pt}
	\begin{subfigure}[b]{0.27\textwidth}
		\centering
		\includegraphics[width=1\textwidth]{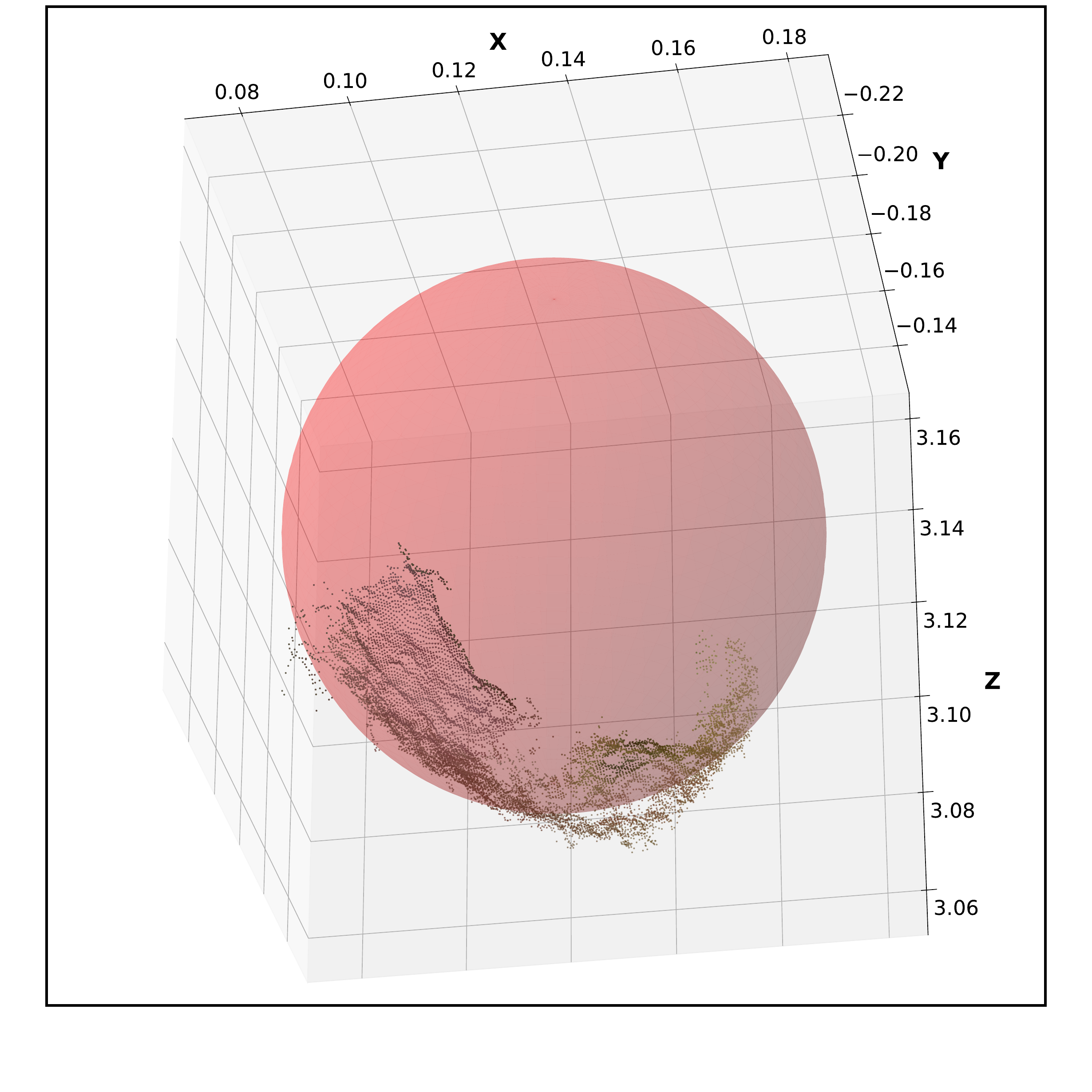}
            % \vspace{-30pt}
		\caption{3D-RANSAC}
		\label{fig_3d_ransac}
	\end{subfigure}
    \caption{The schematics for size estimation algorithms. In (a)-(c) the size estimation is based on the features on the 2D images whereas (d)-(f) are based on the 3D point-cloud of each apple.}
    \label{fig_sizing}
\end{figure}

We compare the following three 3D-based size estimation algorithms:

\begin{enumerate}
\item Largest Segment (\textbf{3D-LSeg}, Fig.~\ref{fig_3d_lseg}): Let $\textbf{P}=\{p_i\}$ denote the apple point cloud, where $p_i=(x_i,y_i,z_i)$. The apple’s diameter is estimated by $d_r=\max\limits_{i,j}||p_i-p_j||_2,$ which is the length of the largest segment within the point cloud.
\item Least Square Fitting (\textbf{3D-LSq}, Fig.~\ref{fig_3d_lsq}): Here, we fit a sphere to the point cloud by minimizing the spherical cost function, $J(\theta)=\sum\limits_i[(x_i-x_c)^2+(y_i-y_c)^2+(z_i-z_c)^2-r^2]^2$, where $(x_i,y_i,z_i)$ is the i-th point from the input point cloud, and $\theta=(x_c,y_c,z_c,r)$ represents the center and radius of the sphere. After solving this optimization problem, the diameter is given by $d_r=2r$
\item RANdom SAmple Consensus (\textbf{3D-RANSAC}, Fig.~\ref{fig_3d_ransac}): We randomly sample four points at a time to define a sphere $f(x_c,y_c,z_c,r)$. We then count how many points in the cloud (inliers) satisfy $-\delta<||p_i-p_c||_2-r<\delta$, where $p_i=(x_i,y_i,z_i)$ is the i-th point in the input point cloud, $p_c=(x_c,y_c,z_c)$ is the estimated apple center position. If the inlier ratio exceeds a certain threshold, the process terminates early; otherwise, it continues for $n$ iterations. As with least-squares fitting,  $d_r=2r$.
\end{enumerate}

\section{Experiment Results} \label{results}
This section presents three main experiments. First, we compare the performance of Grounding-DINO against other state-of-the-art fruit detection models. Second, we evaluate various size estimation algorithms on the entire dataset using ground-truth bounding boxes as input. Finally, we test how well these size estimation algorithms perform when the inputs come from our detection model, thereby analyzing whether detection accuracy influences the size estimation results.

All detection models were trained on a server running Ubuntu 20.04, equipped with an AMD Ryzen Threadripper 3990X 64-Core CPU, two Quadro RTX 8000 GPUs, and 256~GiB of RAM. The detection and size estimation tests were conducted on an Alienware laptop running Ubuntu 22.04, featuring an Intel\textsuperscript{\textregistered} Core\textsuperscript{TM} i9-13900HX CPU, an NVIDIA GeForce RTX 4090 GPU, and 64~GiB of RAM.

\subsection{Detection Model Evaluation} \label{ripeness_eval}
We evaluated the performance of Grounding-DINO on our Fuji ripeness dataset and compared it against other leading detection models, including Faster-RCNN, YOLOv8, Deformable DETR, and DINO. To measure detection performance, we used mean average precision (mAP) and mean average recall (mAR). We also measured the frame rate (FPS) to assess runtime speed.

All models were trained using approximately 75\% of the dataset (the training subset) and then validated on the remaining 25\% (the validation subset). Table~\ref{detection_results} shows that Grounding-DINO with a Swin-B-Transformer backbone achieves the highest mAP and mAR, indicating superior ability to detect ripe and unripe apples. By incorporating text information during training, Grounding-DINO effectively learns both the similarities and differences between ripe and unripe apples, surpassing models that rely only on visual features. Note that Grounding-DINO with the Swin-B-Transformer backbone also achieves an FPS of about 68.3, making it suitable for real-time operations.

\begin{table}[b]
\centering
\begin{tabular}{|c|cccc|c|}
\hline 
Model & mAP50 & mAP75 & mAP50:95 & mAR & FPS \\
\hline \hline
Faster-RCNN (ResNet18) & 78.2 & 68.8 & 59.7 & 74.1 & 70.5 \\
Faster-RCNN (ResNet50) & 77.9 & 68.9 & 59.1 & 74.4 & 70.8 \\
Faster-RCNN (ResNet101) & 77.5 & 71.7 & 62.9 & 76.7 & 70.1 \\
% \hline
% RetinaNet (ResNet18) & 73.3 & 61.8 & 54.8 & & 70.3 \\
YOLOv8s & 75.9 & 68.7 & 61.6 & 82.6 & \textbf{119.3} \\
YOLOv8m & 78.1 & 72.3 & 64.8 & 85.1 & 69.7 \\
YOLOv8l & 78.6 & 72.9 & 65.1 & 85.0 & 95.4 \\
VitDet & 82.3 & 74.6 & 64.7 & 78.7 & 69.8 \\
Deformable-DETR & 73.5 & 65.2 & 59.1 & 79.0 & 69.6 \\
DINO & 79.3 & 72.4 & 65.8 & 87.7 & 71.0 \\
\hline
Grounding-DINO (ResNet) & 72.3 & 65.4 & 59.2 & 85.1 & 68.1 \\
\textbf{Grounding-DINO (Swin-B)} & \textbf{85.9} & \textbf{80.6} & \textbf{72.8} & \textbf{89.6} & 68.3 \\
\hline 
\end{tabular}
\caption{Evaluation on Detection Models, the Grounding-DINO with Swin-B as backbone achieves the highest mean average precision as well as mean average recall among all the models. As for the FPS, the Grounding DINO is running at a speed comparable to most of the other models.}
\label{detection_results}
\end{table}

\begin{figure}[!h]
    \centering
    \includegraphics[width=0.8\textwidth]{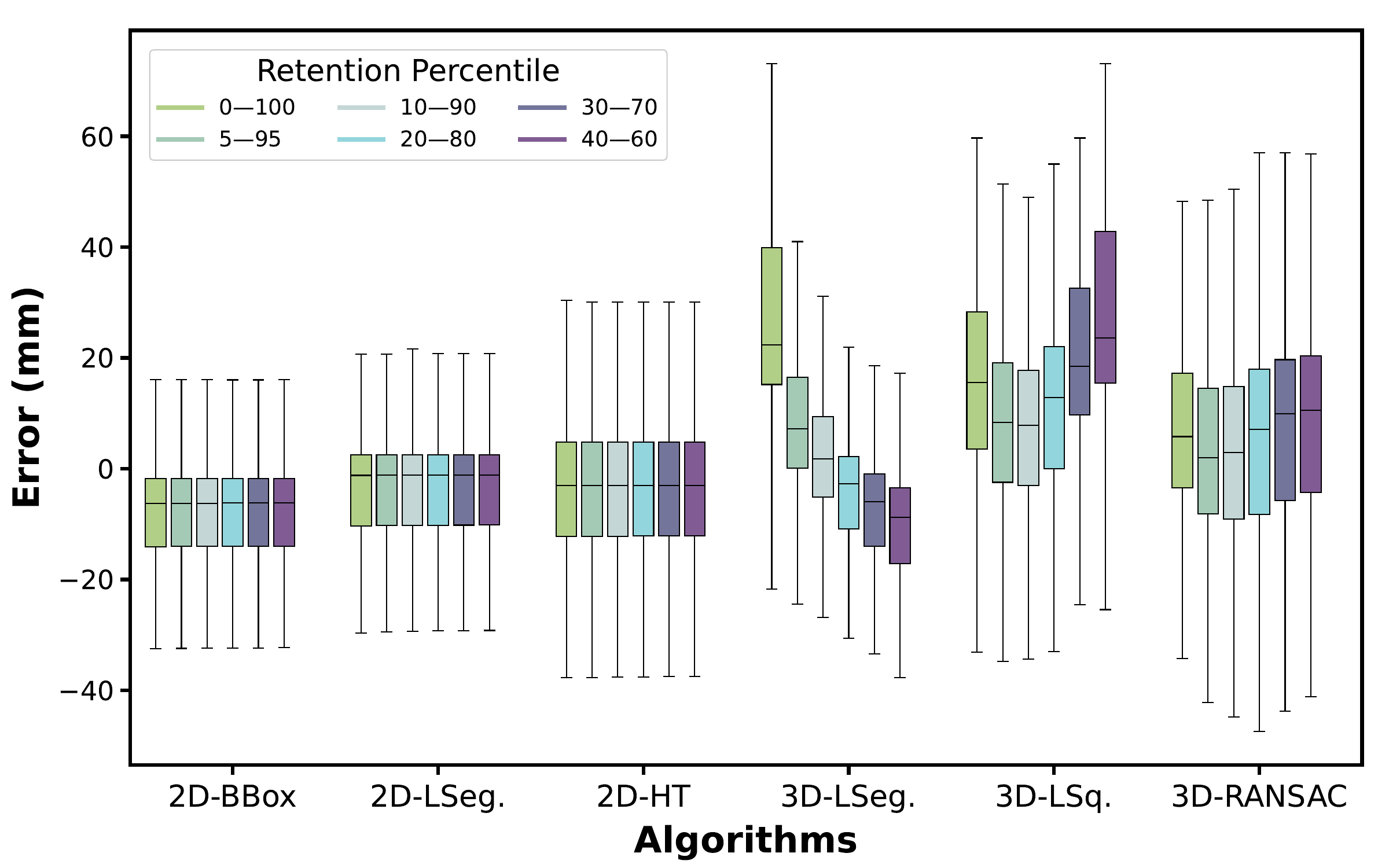}
    \caption{The size estimation results with ground-truth bounding boxes. For each box plot, the upper edge of the box represents 75\% percentile of the data (denote as $Q_3$), while the lower edge represents 25\% ($Q_1$). The black line ($Q_2$) in the middle of the box is the 50\% percentile (median) of the data. The upper whisker $Q_4=Q_3+1.5(Q_3-Q_1)$, while the lower whisker $Q_0=Q_1-1.5(Q_3-Q_1)$. Data points outside $Q_0$ and $Q_4$ are considered outliers, which are omitted in the plot.
    The 2D-based algorithms have less error than 3D, and they are also more robust to the change of outlier removal rates, indicating that they are more capable of tolerating the noise in depth measurement.}
    \label{fig_sizing_gt_res}
\end{figure}
	
\subsection{Size Estimation with Ground-Truth Bounding Boxes} \label{size_estimation_gt}
We first evaluated each size estimation algorithm (from Section~\ref{sec_sizing}) on the entire dataset, using ground-truth bounding boxes as input. The results, shown in Fig.~\ref{fig_sizing_gt_res}, are presented as box plots grouped by the size estimation algorithms. Within each group, we tested different outlier-removal percentiles, ranging from 0\%--100\% to 40\%--60\% retention rate. The y-axis in each plot shows the estimation error, defined as $d_{est}-d_{gt}$, where $d_{est}$ is the estimated diameter of the apple, and $d_{gt}$ is the ground-truth diameter. 

Fig.~\ref{fig_sizing_gt_res} shows the box plots for each algorithms under different retention percentiles. From the figure, we see that 2D-LSeg exhibits the smallest error, with a median value closest to zero. Its interquartile range is also relatively narrow, indicating consistent performance throughout the dataset. Additionally, 2D-LSeg is less sensitive to changes in outlier-removal percentiles, as it only depends on the average depth of apple pixels for size estimation. In contrast, the fitting-based algorithms (3D-LSq, 3D-RANSAC, and 2D-Hough) produce larger errors overall. A key factor is that apples are not perfectly spherical; occlusions and irregular shapes can lead to smaller sphere fits, reducing accuracy.

We also observe that 3D-based algorithms tend to have larger errors and greater sensitivity to outlier-removal settings. Differences in how the retention percentile is chosen can significantly influence the distribution of errors. Besides the issues of occlusion and irregular fruit shapes, depth measurements can be noisier than RGB channels, which further affects the accuracy of 3D-based estimations.

\subsection{Size Estimation After Detection} \label{size_estimation_det}
We next tested size estimation algorithms on apples detected by our trained model, to assess whether detection accuracy affects size estimation. Specifically, we ran the detection model on images in the test set and matched each detected bounding box to a ground-truth bounding box if their Intersection over Union (IOU) exceeded 70\%. The diameter in the corresponding ground-truth bounding box then served as the ground-truth size.

\begin{figure}[!h]
    \centering
    \includegraphics[width=0.8\textwidth]{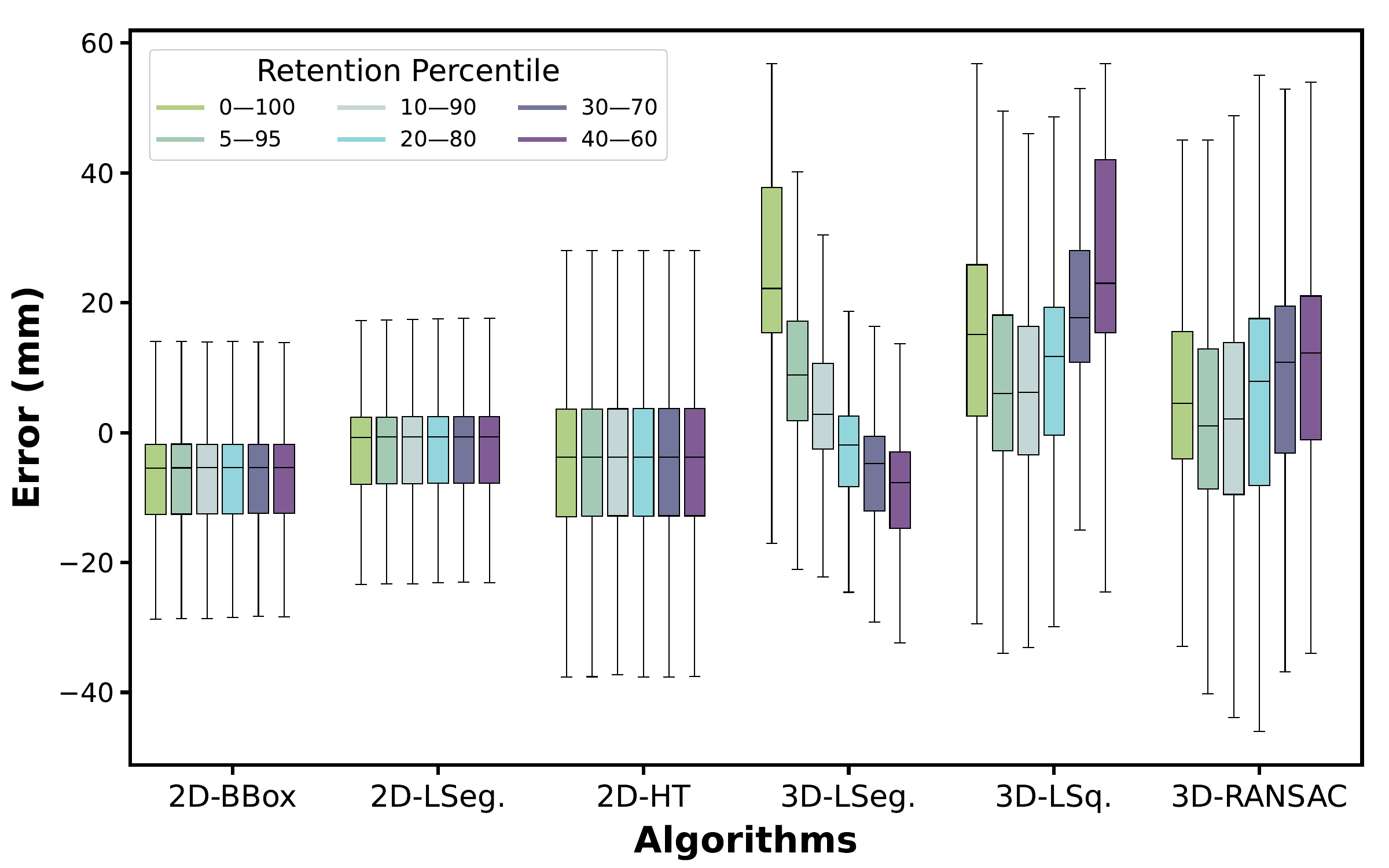}
    \caption{The size estimation results with detected bounding boxes. The result is similar to the result on ground-truth bounding-boxes.}
    \label{fig_sizing_detected_res}
\end{figure}

Figure~\ref{fig_sizing_detected_res} shows the results, which closely resemble those in Fig.~\ref{fig_sizing_gt_res}. These similarities suggest that detection quality does not significantly alter the accuracy of size estimation, indicating robust performance by our proposed approach.

\section{Discussions}
Compared to fruit detection, determining fruit ripeness is more challenging and has received relatively little attention. Two main difficulties hinder progress: (1) collecting appropriate datasets is arduous, and (2) variations in ripeness appear much more subtle than the differences between different object classes.

Traditional feature-engineering methods can classify fruit ripeness but often rely on specific parameter choices that reduce their ability to generalize. By contrast, deep-learning approaches learn directly from labeled data, eliminating the need for manual feature selection and often yielding better results. However, typical deep neural networks—especially those based on convolutional architectures—usually require large datasets to perform well. Through the integration of Transformers and language models, our approach offers stronger capability to handle minor visual differences between ripeness classes and to learn from datasets more efficiently. Nevertheless, several limitations remain, which we outline below. We also propose potential future solutions.

\subsection{Real-Time Inference}
Inference time for ripeness detection and size estimation is a critical concern in automated harvesting. During each picking cycle, a robot must identify the apple’s location, plan a trajectory, and execute the pick. Given limited computing resources and communication bandwidth, every stage of the process needs to be as fast as possible.

Grounding-DINO demonstrates excellent frame rates on our workstation, but its performance on lower-cost or embedded hardware remains a concern. Additionally, the Segment-Anything Model used for size estimation also has considerable computational overhead.

A promising solution is to apply model reduction techniques~\citep{choudhary2020comprehensive,zhu2024survey,gou2021knowledge}, which can decrease the number of parameters without sacrificing much performance. Examples include pruning and knowledge distillation. By compressing the underlying networks, we can achieve faster estimation of fruit ripeness and size with lower memory and processing demands.

\subsection{Ripeness Labeling}
Labeling fruit ripeness presents many challenges, especially for apples in an ``intermediate'' stage. Expert knowledge is usually required to accurately classify these cases and develop a high-quality dataset. By contrast, identifying apples that are distinctly unripe or clearly ripe is relatively straightforward.

A valuable strategy involves label-efficient learning (a form of semi-supervised learning). Consider a dataset where (1) fully unripe and fully ripe apples are labeled, and (2) only a few intermediate cases are annotated. A model trained via label-efficient methods can then infer labels for the remaining intermediate apples, refining its parameters in the process.

Further improvements may come from integrating language models. If we carefully design label names, a language-infused, label-efficient approach could facilitate both high-accuracy ripeness detection and further annotation of partially labeled datasets.

\subsection{Sizing Under Canopy Occlusion}
Although our experiments show that the maximum distance measured in the 2D mask (2D-LSeg) yields the smallest error and variance, it can fail under heavy occlusion. Some studies propose generating an amodal mask that includes both visible and occluded regions~\citep{gene2024amodalapplesize_rgb}, but the limited availability of high-quality amodal annotations for apples remains a barrier. Recent computer vision methods could potentially use existing amodal datasets (even if they are not apple-specific) to train a more general amodal segmentation model. With better estimations of occluded regions, the 2D-max-distance algorithm could become more robust.

Another promising path is to develop an end-to-end framework for size estimation. Rather than splitting the pipeline into multiple stages (e.g., segmentation, mask generation, point cloud generation), a single network could directly learn the relationship between apple images and the corresponding size labels, much like how Faster-RCNN streamlined object detection into a single framework.

\section{Conclusions} \label{conclusion}
This paper focused on evaluating the ``harvestability'' of Fuji apples based on two key factors: ripeness and size, which is essential to determine when a fruit is ready to be picked. By selectively harvesting only those fruits that meet certain criteria, growers can maximize revenue. 
Specifically, for ripeness estimation, we re-labeled two existing Fuji apple datasets to include ripeness annotations. We then trained and tested Grounding-DINO, a cutting-edge object detector that leverages language inputs, to classify Fuji apples as ripe or unripe based on appearance. Our experiments showed that Grounding-DINO outperformed popular detection models by a clear margin.
For size estimation, we leveraged ground-truth size measurements in our labeled Fuji datasets and compared six different algorithms. The 2D-max-distance method showed its robustness to the noise in the depth images, and delivers the smallest error, making it the most robust choice among the tested techniques. 
Both the re-labeled dataset and all relevant source code are publicly available for the research community. This contribution will spur further development of automated selective harvesting systems and encourage future research into more advanced ripeness assessment and size estimation methods.

Despite these advancements, some limitations remain, pointing to potential areas for future research. First, developing an in-field dataset specifically for apple ripeness, with a more rigorous yet efficient annotation process, could enhance ripeness estimation. Second, designing a lightweight yet powerful detection/segmentation model would improve both onboard implementation and subsequent size estimation. Third, making the size estimation algorithm more robust to occlusion—common in harvesting scenarios—would increase its practical applicability.

\section*{Acknowledgment}
This project was funded by the USDA-SCRI project "AIMS for Apple Harvest and In-Field Sorting" (Project No. 2023-51181-41244) and the NSF project "NRI: INT: SMART: Soft Multi-Arm RoboT for Synergistic Collaboration with Humans" (Project No. NSF ECCS-2024649). 

% \section*{Authorship Contribution}
% \textbf{Jiajia Li}: Conceptualization, Investigation, Software, Writing – original draft; \textbf{Kyle Lammers}: Conceptualization, Investigation, Writing – original draft; \textbf{Jiajia Li}:Investigation, Software, Writing – original draft; \textbf{Zhaojian Li}: Supervision, Writing – review.

% \section*{Acknowledgement}
% This work was supported in part by XX

\typeout{}
\bibliography{ref}
\end{document}